\pdfoutput=1

\documentclass[11pt]{article}

\usepackage[]{acl}

\usepackage{times}
\usepackage{latexsym}

\usepackage[T1]{fontenc}

\usepackage[utf8]{inputenc}

\usepackage{microtype}

\usepackage[inline]{enumitem}

\usepackage{graphicx}
\usepackage{epstopdf}
\usepackage{booktabs}
\usepackage{amsthm,amsmath,amssymb,amssymb,bm,mathrsfs}
\usepackage{textcomp}
\usepackage{xcolor}
\usepackage{changes}
\usepackage{algorithm}
\usepackage{ulem}
\usepackage{subcaption}
\usepackage{multirow}
\usepackage{ulem}

\title{Is Neural Topic Modelling Better than Clustering? An Empirical Study on Clustering with Contextual Embeddings for Topics}

\author{Zihan Zhang\textsuperscript{1},  Meng Fang\textsuperscript{2}, Ling Chen\textsuperscript{1}, Mohammad-Reza Namazi-Rad\textsuperscript{3} \\
        \textsuperscript{1}AAII, University of Technology Sydney, NSW, Australia \\
        \texttt{Zihan.Zhang-5@student.uts.edu.au}, \texttt{Ling.Chen@uts.edu.au} \\
        \textsuperscript{2}Eindhoven University of Technology, Eindhoven, the Netherlands \\
        \texttt{m.fang@tue.nl}\\
        \textsuperscript{3}NIASRA, University of Wollongong, NSW, Australia \\
        \texttt{mrad@uow.edu.au} \\
        }

\begin{document}
\maketitle
\begin{abstract}
Recent work incorporates pre-trained word embeddings such as BERT embeddings into Neural Topic Models (NTMs), generating highly coherent topics. However, with high-quality contextualized document representations, do we really need sophisticated neural models to obtain coherent and interpretable topics? In this paper, we conduct thorough experiments showing that directly clustering high-quality sentence embeddings with an appropriate word selecting method can generate more coherent and diverse topics than NTMs, achieving also higher efficiency and simplicity.\footnote{Code is available at \url{https://github.com/hyintell/topicx}}

\end{abstract}

\section{Introduction}

Topic modelling is an unsupervised method to uncover latent semantic themes among documents \citep{boyd2017applications}. Neural topic models (NTMs) \citep{miao2016neural, srivastava2017autoencoding} incorporating neural components have significantly advanced the modelling results than the traditional Latent Dirichlet Allocation (LDA; \citealt{blei2003latent}).
Later, contextualized word and sentence embeddings produced by pre-trained language models such as BERT \citep{devlin-etal-2019-bert} have demonstrated the  state-of-the-art results in multiple Natural Language Processing (NLP) tasks \citep{xia-etal-2020-bert}, which attracts attentions from the topic modelling community. 
Recent work has successfully incorporated these contextualized embeddings into NTMs, showing improved topic coherence than conventional NTMs that use Bag-of-Words (BoW) as document representations \citep{ bianchi-etal-2021-pre, bianchi-etal-2021-cross, jin-etal-2021-neural}.
Despite the promising performance, existing NTMs are generally based on a variational autoencoder framework (VAE; \citealt{kingma2013auto}), which suffers from hyper-parameters tuning and computational overheads \citep{zhao2021topic}. Moreover, the integration of the pre-trained embeddings to the standard VAE framework adds additional model complexity. With high-quality contextualized document representations, do we really need sophisticated NTMs to obtain coherent and interpretable topics?






Recent work \citep{aharoni-goldberg-2020-unsupervised, sia-etal-2020-tired, thompson2020topic, grootendorst2020bertopic} has shown that directly congregating contextualized embeddings can get semantically similar word or document clusters. Specifically, \citet{sia-etal-2020-tired} cluster \textit{vocabulary-level} word embeddings and obtain top words from each cluster using weighing and re-ranking, while \citet{thompson2020topic} consider polysemy and perform \textit{token-level} clustering. However, the use of term frequency (TF) to select topic words fails to capture the semantics of clusters precisely because words with high frequency may be common across different clusters. \citet{grootendorst2020bertopic} propose a class-based Term Frequency-
Inverse Document Frequenc (c-TF-IDF) method that extract important words from each clustered documents, which tends to choose representative words within each cluster to form topics. However, it overlooks the global semantics between clusters which could be incorporated. In addition, all above works only compare the performance with the traditional LDA while ignoring the promising NTMs proposed recently. The performance of the clustering-based topic models is still yet uncovered.






\textit{Is neural topic modelling better than simple embedding clustering}?
This work compares the performance of NTMs and contextualized embedding-based  clustering systematically. Our main focus is to provide insights by comparing the two paradigms for topic models, which has not been investigated before. We employ a straightforward framework for clustering. In addition, we explore different  strategies to select topic words for clusters. We evaluate our approach on three datasets with various text lengths.

Our contributions are as follows:
First, we find that directly clustering high-quality sentence embeddings can generate as good topics as NTMs, providing a simple and efficient solution to uncover latent topics among documents. Second, we propose a new topic word selecting method, which is the key to producing highly  coherent and diverse topics. Third, we show that the clustering-based model is robust to the length of documents and the number of topics. Reducing the embedding dimensionality negligibly affects the performance but saves runtime. From our best knowledge, we are the first to compare with NTMs, using contextualized embeddings that produced by various transformer-based models.

\section{Models}
This study compares embedding clustering-based models with LDA and a series of existing NTMs as follows. Implementation details are supplied in Appendix \ref{appendix_implementation_details}.

\textbf{LDA} \citep{blei2003latent}: the representative traditional topic model in history that generates topics via document-topics and topic-words distributions.

\textbf{ProdLDA} \citep{srivastava2017autoencoding}: a prominent NTM that employs the VAE \citep{kingma2013auto} to reconstruct the BoW representation.

\textbf{CombinedTM} \citep{bianchi-etal-2021-pre}:  extends ProdLDA by concatenating the contextualized SBERT \citep{reimers-gurevych-2019-sentence} embeddings with the original BoW as the new input to feed into the VAE framework.

\textbf{ZeroShotTM} \citep{bianchi-etal-2021-cross}: also builds upon ProdLDA, but it replaces the original BoW with SBERT embeddings entirely.

\textbf{BERT+KM} \citep{sia-etal-2020-tired}: a clustering-based method that first uses K-Means (KM) to cluster word embeddings, then apply TF to weight and re-rank words to obtain topic words. 

\textbf{BERT+UMAP+HDBSCAN} (i.e., BERTopic) \citep{grootendorst2020bertopic}: a clustering-based method that first leverages HDBSCAN \citep{mcinnes2017accelerated} to cluster BERT embeddings of the sentences and Uniform Manifold Approximation Projection (UMAP) \citep{mcinnes2018umap} to reduce embedding dimensions, then use a class-based TFIDF (i.e. c-TF-IDF) to select topic words within each cluster. Note that BERTopic may not generate the specified number of topics.

\begin{figure}
  \centering
  \includegraphics[width=\columnwidth]{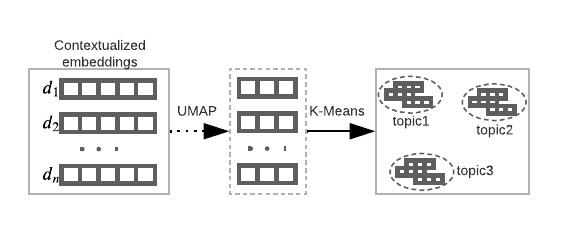}
  \caption{\footnotesize Architecture of our method. Reducing embedding dimension is optional but can save runtime (see Section \ref{ablation_studies}).}
  \label{model_architecture}
\end{figure}

\textbf{Contextual Embeddings+UMAP+KM} (our method CETopic):
we use a simple clustering framework with contextualized embeddings for topic modelling, as shown in Figure \ref{model_architecture}.
We first encode pre-processed documents to obtain contextualized sentence embeddings through pre-trained language models. After that, we lower the dimension of the embeddings before applying clustering methods (e.g., K-Means; KM) to group similar documents. Each cluster will be regarded as a topic. Finally, we adopt a weighting method to select representative words as topics. 


We believe that high-quality document embeddings are critical for clustering-based topic modelling. We thus experiment with different embeddings including BERT, RoBERTa \citep{liu2019roberta}, and SBERT. We also adopt SimCSE \citep{gao-etal-2021-simcse}, a recently proposed sentence embeddings of contrastive learning, that has shown the state-of-the-art performance on multiple semantic textual similarity tasks. Both supervised and unsupervised SimCSE are investigated in our experiment (e.g., Table~\ref{complete_results}).



Pre-trained contextualized sentence embeddings often have high dimensionalities. To reduce the computational cost, we apply the UMAP in our implementation to reduce the  dimensionality while maintaining the essential information of the embeddings. We find that reducing dimensionality before clustering has a negligible impact on performance (Section \ref{ablation_studies}).

We cluster the dimension-reduced sentence embeddings using K-Means because of its efficiency and simplicity. Semantically close documents are gathered together, and each cluster is supposed to represent a topic. 


\section{Topic Words for Clusters}
Once we have a group of clustered documents, selecting representative topic words is vital to identify semantics of topics. 
Inspired by previous works \citep{ramos2003using, grootendorst2020bertopic}, we explore several weighting metrics to obtain topic words in clusters. Let $n_{t,d}$ be the frequency of word $t$ in document $d$, $\sum_{t^{\prime}} n_{t^{\prime},d}$ be the total words' frequency in the document, and $D$ be the entire corpus. Term Frequency-Inverse Document Frequency (TFIDF) is defined as $    \mathbf{TFIDF}=\frac{n_{t,d}}{\sum_{t^{\prime}} n_{t^{\prime},d}} \cdot \log \left(\frac{|D|}{|\{d \in D : t \in d\}|}\right)$.
While capturing the word importance across the entire corpus, TFIDF ignores that semantically similar documents have been grouped together. To address this issue, we consider two alternative strategies. First, we concatenate the documents within a cluster to be a single long document and calculate the term frequency of each word in each cluster:
\begin{equation}
\small
    \mathbf{TF_{i}}=\frac{n_{t,i}}{\sum_{t^{\prime}} n_{t^{\prime},i}}
\end{equation}
where $n_{t,i}$ is the frequency of word $t$ in cluster $i$, $\sum_{t^{\prime}} n_{t^{\prime},i}$ is the total word frequency in the cluster. Second, for each cluster $i$, we apply TFIDF:
\begin{equation}
\small
    \mathbf{TFIDF_{i}}=\frac{{n_{t,d}}_{i}}{\sum_{t^{\prime}} {n_{t^{\prime},d}}_{i}} \cdot \log \left(\frac{|D_{i}|}{|\{d \in D_{i} : t \in d\}|}\right)
\end{equation}
where ${n_{t,d}}_{i}$ denotes the frequency of word $t$ in document $d$, which is in cluster $i$, and $|D_{i}|$ is the number of documents in cluster $i$.

Besides the two local cluster-based strategies, we further incorporate the global word importance with local term frequency within each cluster: 
\begin{equation}
\small
    \mathbf{TFIDF \times TF_{i}}=\mathbf{TFIDF} \cdot \mathbf{TF_{i}}
\end{equation}
and we combine the global word importance with term frequency across clusters:
\begin{equation}
\small
    \mathbf{TFIDF \times IDF_{i}}=\mathbf{TFIDF} \cdot \log \left(\frac{|K|}{|\{t \in K\}|}\right) 
    \label{eq_tfidf_idfi}
\end{equation}
where $|K|$ is the number of clusters and $|\{t \in K\}|$ is the number of clusters that word $t$ appears. 


\section{Experiments}

\subsection{Datasets}

We adopt three datasets of various text lengths in our experiments, namely 20Newsgroups\footnote{\url{http://qwone.com/~jason/20Newsgroups/}}, M10 \citep{pmlr-v39-lim14}, and BBC News \citep{greene2006practical}. We follow OCTIS \citep{terragni-etal-2021-octis} to pre-process these raw datasets. The statistics of the datasets are shown in Table \ref{datasets-statistics}.

\begin{table}[H]
\centering
\resizebox{0.4\textwidth}{!}{
\begin{tabular}{l|rrrr}
\hline
\textbf{Dataset} & \textbf{$D$} & \textbf{$V$} & \textbf{$L$} & \textbf{$N_d$} \\
\hline
20Newsgroups & 16,309 & 1,612 & 20 & 48 \\
M10 & 8,355 & 1,696 & 10 & 5.9 \\
BBC News & 2,225 & 2,949 & 5 & 120 \\
\hline
\end{tabular}
}
\caption{\label{datasets-statistics}
Statistics of the pre-processed datasets, where $D$ denotes the total number of documents, $V$ denotes the vocabulary size, $L$ denotes the number of corpus categories, and $N_d$ denotes the average number of words per document. 
}
\end{table}

\begin{table*}
\centering
\small
\addtolength{\tabcolsep}{-1.5pt}
\begin{tabular}{l|ccc|ccc|ccc}
\hline
& \multicolumn{3}{c|}{\textbf{20Newsgroups}} & \multicolumn{3}{c|} {\textbf{M10}} & \multicolumn{3}{c}{\textbf{BBC News}} \\
\hline
\textbf{Model} & \textit{TU} & \textit{NPMI} & \textit{$C_V$} & \textit{TU}  & \textit{NPMI} & \textit{$C_V$} & \textit{TU} & \textit{NPMI}  & \textit{$C_V$} \\
\hline
LDA & 0.717 & 0.040 & 0.511 & 0.681 & -0.177 & 0.336 & 0.312 & -0.014 & 0.357 \\
ProdLDA & 0.736 & 0.045 & 0.574 & 0.650 & -0.260 & 0.432 & 0.702 & -0.044 & 0.540 \\
CombinedTM & 0.700 & 0.065 & 0.601 & 0.581 & 0.001 & 0.443 & 0.606 & 0.042 & 0.639 \\ 
ZeroShotTM & 0.729 & 0.069 & 0.614 & 0.633 & -0.056 & 0.433 & 0.699 & -0.050 & 0.531 \\ 
\hline
BERT$_{\texttt{base}}$+KM$^\dagger$ & 0.346 & 0.065& 0.521 & 0.484 & 0.116 & 0.588 &  0.529 & 0.111 & 0.637 \\
BERT$_{\texttt{base}}$+UMAP+HDBSCAN$^\ddagger$ & \textbf{0.805} & 0.059 & 0.534 & 0.730 & -0.017 & 0.606 &  0.732 & 0.089 & 0.686 \\
\hline
BERT$_{\texttt{base}}$$^*$ & 0.562 & 0.118 & 0.649 & 0.763 & 0.146 & 0.725 & 0.689 & 0.129 & 0.700 \\
BERT$_{\texttt{large}}$$^*$ & 0.550 & 0.116 & 0.646 & 0.743 & 0.138 & 0.715 & 0.684 & 0.132 & 0.705 \\
RoBERTa$_{\texttt{base}}$$^*$ & 0.385 & 0.028 & 0.464 & 0.634 & -0.008 & 0.480 & 0.671 & 0.098 & 0.646 \\
RoBERTa$_{\texttt{large}}$$^*$ & 0.404 & 0.014 & 0.440 & 0.669 & 0.001 & 0.506 & 0.673 & 0.046 & 0.555 \\
\hline

BERT$_{\texttt{base}}$+UMAP$^*$ & 0.589 & 0.128 & 0.671 & 0.794 & 0.159 & 0.706 & 0.716 & 0.135 & 0.716 \\
BERT$_{\texttt{large}}$+UMAP$^*$ & 0.563 & 0.126 & 0.662 & 0.751 & 0.176 & 0.681 & 0.721 & 0.139 & 0.720 \\
RoBERTa$_{\texttt{base}}$+UMAP$^*$ & 0.434 & 0.063 & 0.522 & 0.640 & 0.091 & 0.547 & 0.710 & 0.106 & 0.664 \\
RoBERTa$_{\texttt{large}}$+UMAP$^*$ & 0.463 & 0.054 & 0.499 & 0.636 & 0.046 & 0.513 & 0.706 & 0.077 & 0.632 \\
\hline
SBERT$_{\texttt{base}}$$^*$ & 0.668 & 0.126 & 0.658 & 0.832 & 0.164 & 0.742 & 0.727 & 0.137 & 0.719 \\
SBERT$_{\texttt{large}}$$^*$ & 0.674 & 0.135 & 0.673 & 0.844 & 0.168 & 0.752 & 0.718 & 0.134 & 0.714 \\
SRoBERTa$_{\texttt{base}}$$^*$ & 0.670 & 0.128 & 0.654 & 0.815 & 0.149 & 0.713 & 0.719 & 0.131 & 0.699 \\
SRoBERTa$_{\texttt{large}}$$^*$ & 0.649 & 0.115 & 0.640 & 0.823 & 0.155 & 0.735 & 0.696 & 0.122 & 0.694 \\
\hline
SBERT$_{\texttt{base}}$+UMAP$^*$ & 0.679 & 0.139 & 0.690 & 0.841 & 0.192 & 0.715 & 0.749 & 0.142 & \textbf{0.730} \\
SBERT$_{\texttt{large}}$+UMAP$^*$ & 0.681 & 0.139 & 0.691 & 0.836 & 0.203 & 0.723 & 0.744 & 0.136 & 0.725 \\
SRoBERTa$_{\texttt{base}}$+UMAP$^*$ & 0.680 & 0.138 & 0.684 & 0.830 & 0.192 & 0.722 & 0.747 & 0.135 & 0.716 \\
SRoBERTa$_{\texttt{large}}$+UMAP$^*$ & 0.680 & 0.131 & 0.670 & 0.799 & 0.196 & 0.700 & 0.728 & 0.121 & 0.705 \\
\hline
Unsup-SimCSE(BERT$_{\texttt{base}}$)$^*$ & 0.677 & 0.147 & 0.694 & 0.831 & 0.180 & 0.750 & 0.730 & 0.142 & 0.722 \\ 
Unsup-SimCSE(BERT$_{\texttt{large}}$)$^*$ & 0.700 & 0.145 & 0.693 & 0.832 & 0.182 & 0.750 & 0.728 & 0.135 & 0.714 \\ 
Unsup-SimCSE(RoBERTa$_{\texttt{base}}$)$^*$ & 0.696 & 0.142 & 0.682 & 0.823 & 0.164 & 0.726 & 0.731 & 0.137 & 0.700 \\ 
Unsup-SimCSE(RoBERTa$_{\texttt{large}}$)$^*$ & 0.722 & 0.147 & 0.694 & 0.812 & 0.171 & 0.734 & 0.736 & 0.142 & 0.711 \\ 
\hline
Unsup-SimCSE(BERT$_{\texttt{base}}$)+UMAP$^*$ & 0.692 & 0.139 & 0.685 & \textbf{0.851} & \textbf{0.206} & 0.744 & 0.733 & 0.146 & 0.729 \\ 
Unsup-SimCSE(BERT$_{\texttt{large}}$)+UMAP$^*$ & 0.694 & 0.145 & 0.698 & 0.843 & 0.200 & 0.721 & 0.736 & 0.128 & 0.709 \\ 
Unsup-SimCSE(RoBERTa$_{\texttt{base}}$)+UMAP$^*$ & 0.689 & 0.145 & 0.703 & 0.843 & 0.192 & 0.726 & 0.747 & 0.130 & 0.701 \\ 
Unsup-SimCSE(RoBERTa$_{\texttt{large}}$)+UMAP$^*$ & 0.717 & 0.146 & 0.701 & 0.813 & 0.190 & 0.710 & 0.752 & 0.138 & 0.713 \\ 
\hline
Sup-SimCSE(BERT$_{\texttt{base}}$)$^*$ & 0.721 & 0.151 & 0.702 & 0.829 & 0.180 & 0.746 & 0.736 & 0.143 & 0.720 \\ 
Sup-SimCSE(BERT$_{\texttt{large}}$)$^*$ & 0.706 & \textbf{0.155} & \textbf{0.709} & 0.833 & 0.189 & \textbf{0.762} & 0.744 & 0.146 & \textbf{0.730} \\ 
Sup-SimCSE(RoBERTa$_{\texttt{base}}$)$^*$ & 0.718 & 0.145 & 0.693 & 0.829 & 0.170 & 0.734 & 0.738 & 0.140 & 0.715 \\ 
Sup-SimCSE(RoBERTa$_{\texttt{large}}$)$^*$ & 0.716 & 0.148 & 0.696 & 0.826 & 0.179 & 0.742 & 0.751 & \textbf{0.147} & 0.726 \\ 
\hline
Sup-SimCSE(BERT$_{\texttt{base}}$)+UMAP$^*$ & 0.714 & 0.146 & 0.698 & 0.815 & 0.202 & 0.730 & 0.739 & 0.143 & 0.724 \\ 
Sup-SimCSE(BERT$_{\texttt{large}}$)+UMAP$^*$ & 0.721 & 0.150 & 0.704 & 0.834 & \textbf{0.206} & 0.728 & 0.750 & 0.145 & 0.729 \\ 
Sup-SimCSE(RoBERTa$_{\texttt{base}}$)+UMAP$^*$ & 0.709 & 0.144 & 0.700 & 0.822 & 0.195 & 0.711 & 0.752 & 0.142 & 0.723 \\ 
Sup-SimCSE(RoBERTa$_{\texttt{large}}$)+UMAP$^*$ & 0.708 & 0.147 & 0.701 & 0.818 & 0.189 & 0.704 & \textbf{0.754} & 0.145 & 0.725 \\ 
\hline
\end{tabular}
\caption{Topic coherence (\textit{NPMI} and \textit{$C_V$}) and topic diversity (\textit{TU}) of the top 10 words. All results are averaged across the 5 number of topics ($K = \{$ground truth, $25, 50, 75, 100\}$). Each model is averaged over 5 runs. Best results are in bold. $\dagger$: we use the method from \cite{sia-etal-2020-tired}, which uses PCA to reduce embedding dimensionality and TF to select words. $\ddagger$: we use BERTopic \cite{grootendorst2020bertopic} (Note that BERTopic cannot reach the specified topic number, thus may have performance increased). $*$: our method CETopic adopts KM to cluster embeddings and $\mathbf{TFIDF \times IDF_{i}}$ (Eq. \ref{eq_tfidf_idfi}) to select topic words. Dimensionality: $\texttt{base}$: 768, $\texttt{large}$: 1024.   }
\label{complete_results}
\end{table*}

\subsection{Evaluation Metrics}

We evaluate the topic quality in terms of both topic diversity and topic coherence: Topic Diversity (\textit{TU}) \citep{nan-etal-2019-topic} measures the uniqueness of the words across all topics; Normalized Pointwise Mutual Information (\textit{NPMI}) \citep{newman-etal-2010-automatic} measures topic coherence internally using a sliding window to count word co-occurrence patterns; Topic Coherence (\textit{$C_V$}) \citep{roder2015exploring} is a variant of \textit{NPMI} that uses the one-set segmentation to count word co-occurrences and the cosine similarity as the similarity measure.



\begin{table}
\centering
\resizebox{0.5\textwidth}{!}{
\begin{tabular}{l|ccc}
\hline
\textbf{Method} & \textbf{Avg} \textit{TU}  & \textbf{Avg} \textit{NPMI}  & \textbf{Avg} \textit{$C_V$} \\
\hline
$\mathbf{TF_{i}}$ & 0.442 & 0.081 & 0.555  \\
$\mathbf{TFIDF_{i}}$ & 0.508 & 0.110 & 0.626 \\
$\mathbf{TFIDF} \times \mathbf{TF_{i}}$ & 0.438 & 0.078 & 0.551  \\ 
$\mathbf{TFIDF} \times \mathbf{IDF_{i}}$ & \textbf{0.689} & \textbf{0.145} & \textbf{0.702}  \\ 
\hline
\end{tabular}
}
\caption{\footnotesize Comparison between different topic word selecting methods on 20Newsgroups using Unsup-SimCSE(RoBERTa$_{\texttt{base}}$)+UMAP with $K=30$.}
\label{word_selecting_compare}
\end{table}




\begin{figure*}
\centering
\includegraphics[width=\textwidth]{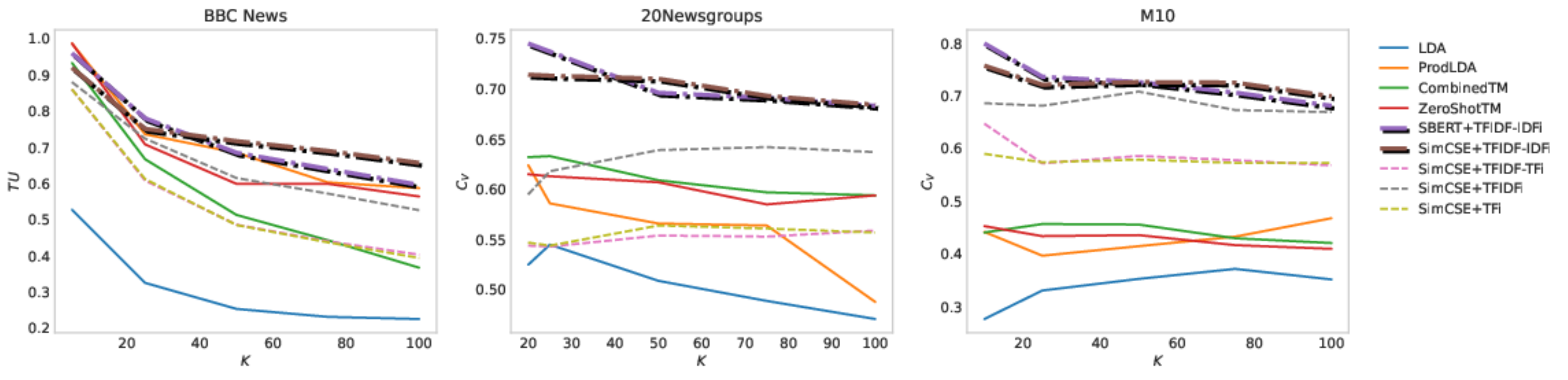}
\caption{\label{fig_different_k} \footnotesize Topic coherence (\textit{$C_V$}) and diversity (\textit{TU}) of different models over different topic number $K$. Cluster models use SBERT$_{\texttt{base}}$+UMAP and Sup-SimCSE(BERT$_{\texttt{base}}$)+UMAP. }
\end{figure*}



\subsection{Results \& Analysis}

We report the main results in Table \ref{complete_results}. 


\textbf{Directly clustering high-quality sentence embeddings can generate good topics.} From Table \ref{complete_results}, it can be observed that SBERT and SimCSE-based clustering models achieve the best averaged topic coherence among the three datasets while maintaining remarkable topic diversities. Conversely, clustering RoBERTa achieves similar or worse results than contextualized NTMs. The results suggest that contextualized embeddings are essential to get high-quality topics.


\textbf{Topic words weighting method is vital.} We can see in Figure \ref{fig_different_k} that inappropriate word selecting methods ($ \mathbf{TFIDF \times TF_{i}}$ and $\mathbf{TF_{i}}$) lead to worse topic coherence than the contextualized NTMs (i.e., CombinedTM and ZeroShotTM), and even the BoW-based ProdLDA. Moreover, from Table \ref{complete_results}, BERT$_{\texttt{base}}$+KM adopt TF to obtain top words for each cluster, which ignores that the words may also be prevalent in other clusters, thus having poor topic diversities. It is also worthy to note that although BERT$_{\texttt{base}}$+UMAP+HDBSCAN (i.e. BERTopic) reaches the highest topic diversity on 20Newsgroups, it cannot produce the specified topic numbers. Thus its performance may be boosted because of the reduced topic numbers. Moreover, our proposed methods, i.e. BERT$_{\texttt{base}}$ and BERT$_{\texttt{base}}$+UMAP outperforms BERTopic in most metrics, especially on topic coherence. This suggests that c-TF-IDF tends to discover incoherent words from each cluster to maintain a high topic uniqueness. Instead, our proposed method, $\mathbf{TFIDF \times IDF_{i}}$, considers the locally important words and globally infrequent words at the same time. We provide more comparison of the word selecting methods in Section \ref{ablation_studies}.

\textbf{Clustering-based topic models are robust to various lengths of documents.} From Table \ref{complete_results} and Figure \ref{fig_different_k}, we find that clustering-based models with high-quality embeddings (SBERT and SimCSE) consistently perform better than conventional LDA and NTMs, especially on the short text dataset M10, even with different word selecting methods.

\subsection{Ablation Studies}
\label{ablation_studies}

We further investigate the impact of the topic word selecting methods, different embedding dimensionalities, as well as the topic numbers.

\textbf{Topic word selecting methods.} Table \ref{word_selecting_compare} shows the comparison between different word weighting methods. $\mathbf{TFIDF \times IDF_{i}}$ achieves significantly better results among all methods. This indicates that $\mathbf{TFIDF}$ marks out the important words to each document in the entire corpus, while $\mathbf{IDF_{i}}$ penalizes the common words in multiple clusters. Conversely, the other three methods ignore that frequent words in a cluster may also be prevalent in other clusters, hence selecting such words leading to low topic diversities. A further analysis in Appendix \ref{compare_topic_words} also supports the observation.

\begin{figure}
  \includegraphics[width=\columnwidth]{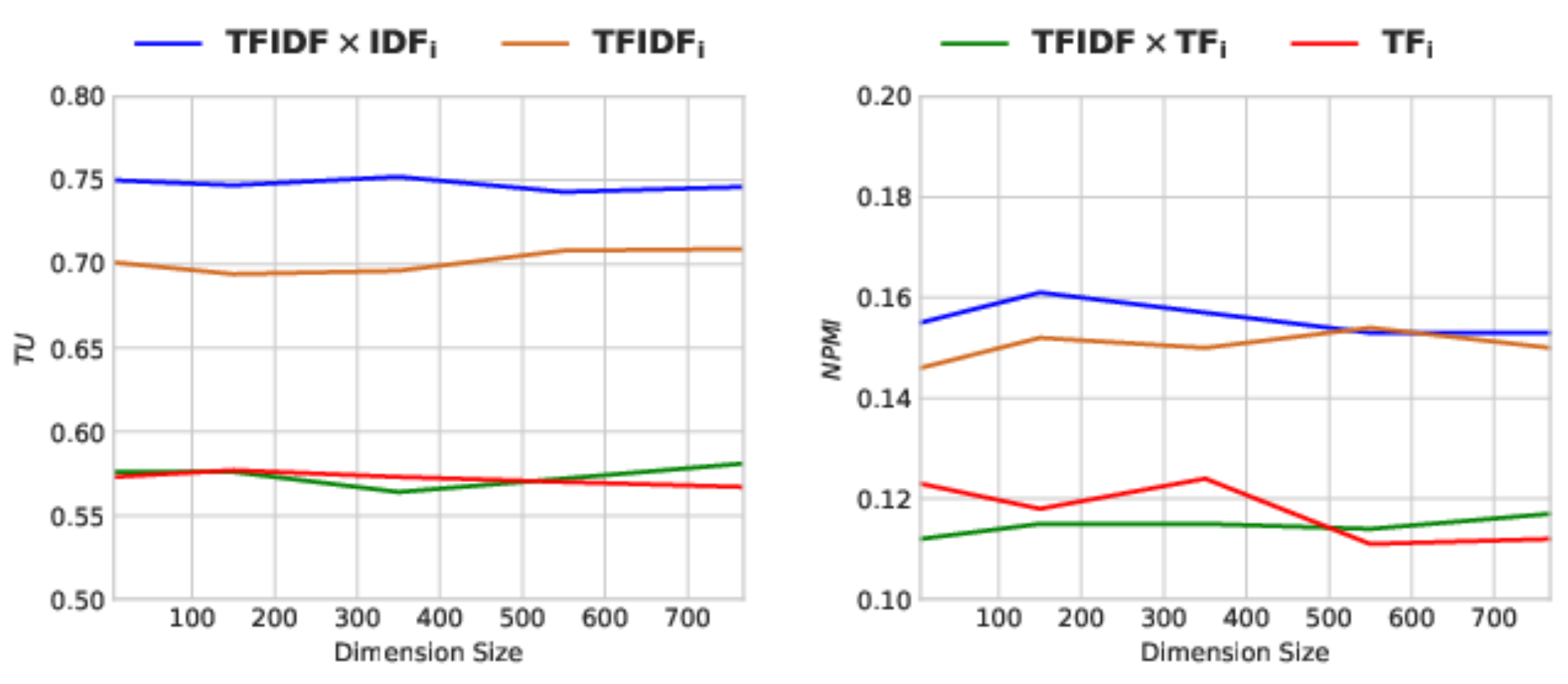}
  \caption{\footnotesize Topic coherence and diversity over different embedding dimensions on BBC News using Unsup-SimCSE(RoBERTa$_{\texttt{base}}$)+UMAP with $K=30$.}
  \label{embedding_dimension}
\end{figure}

\textbf{Embedding dimensionality reduction.} We apply UMAP to reduce the dimensionality of the sentence embeddings before clustering. As shown in Figure \ref{embedding_dimension}, the  embeddings dimensionality negligibly affects topic quality for all word selecting methods. However, reducing to a lower dimensionality decreases the computational runtime as shown in Table \ref{runtime_compare}. We compare the model runtime between the contextualized NTM CombinedTM and clustering-based models. We reduce the dimensionality of the sentence embeddings to 50 using UMAP. All models run on NVIDIA T4 GPU. 

\begin{table}[H]
\centering
\resizebox{0.5\textwidth}{!}{
\begin{tabular}{c|c}
\hline
\textbf{Model} & \textbf{Runtime} \\
\hline
CombinedTM & 149s \\
SBERT(BERT$_{\texttt{base}}$)  & 113s \\
SBERT(BERT$_{\texttt{base}}$)+UMAP to \texttt{dim}=50 & 101s \\
\hline
\end{tabular}
}
\caption{\footnotesize Runtime comparison on 20Newsgroups with $K = 30$. Results are averaged across 5 runs.}
\label{runtime_compare}
\end{table}

\textbf{Topic numbers $K$.} We investigate the impact of the different number of topics $K$ on the performance of the models. Figure \ref{fig_different_k} plots the trends of \textit{TU} and \textit{$C_V$} on three datasets. We observe that the \textit{TU} of clustering-based topic models, especially the models using $\mathbf{TFIDF \times IDF_{i}}$, decrease slowly compared to others when $K$ increases. 
The similar trend can be observed for topic coherence, while the \textit{$C_V$} of LDA and NTMs either fluctuates significantly or stays at a low level. 






\section{Conclusion}

We conduct a thorough empirical study to show that a clustering-based method can generate commendable topics as long as high-quality contextualized sentence embeddings are used, together  with an appropriate topic word selecting strategy. Compared to neural topic models, clustering-based models are more simple, efficient and robust to various document lengths and topic numbers, which can be applied in some situations as an alternative.


\section*{Acknowledgement}

This work is fully funded by TPG Telecom. We thank anonymous reviewers for their valuable comments. We also thank Yunqiu Xu for discussions.

\bibliography{acl}

\begin{thebibliography}{27}
\expandafter\ifx\csname natexlab\endcsname\relax\def\natexlab#1{#1}\fi

\bibitem[{Aharoni and Goldberg(2020)}]{aharoni-goldberg-2020-unsupervised}
Roee Aharoni and Yoav Goldberg. 2020.
\newblock \href {https://doi.org/10.18653/v1/2020.acl-main.692} {Unsupervised
  domain clusters in pretrained language models}.
\newblock In \emph{Proceedings of the 58th Annual Meeting of the Association
  for Computational Linguistics}, pages 7747--7763, Online. Association for
  Computational Linguistics.

\bibitem[{Bianchi et~al.(2021{\natexlab{a}})Bianchi, Terragni, and
  Hovy}]{bianchi-etal-2021-pre}
Federico Bianchi, Silvia Terragni, and Dirk Hovy. 2021{\natexlab{a}}.
\newblock \href {https://doi.org/10.18653/v1/2021.acl-short.96} {Pre-training
  is a hot topic: Contextualized document embeddings improve topic coherence}.
\newblock In \emph{Proceedings of the 59th Annual Meeting of the Association
  for Computational Linguistics and the 11th International Joint Conference on
  Natural Language Processing (Volume 2: Short Papers)}, pages 759--766,
  Online. Association for Computational Linguistics.

\bibitem[{Bianchi et~al.(2021{\natexlab{b}})Bianchi, Terragni, Hovy, Nozza, and
  Fersini}]{bianchi-etal-2021-cross}
Federico Bianchi, Silvia Terragni, Dirk Hovy, Debora Nozza, and Elisabetta
  Fersini. 2021{\natexlab{b}}.
\newblock \href {https://doi.org/10.18653/v1/2021.eacl-main.143} {Cross-lingual
  contextualized topic models with zero-shot learning}.
\newblock In \emph{Proceedings of the 16th Conference of the European Chapter
  of the Association for Computational Linguistics: Main Volume}, pages
  1676--1683, Online. Association for Computational Linguistics.

\bibitem[{Blei et~al.(2003)Blei, Ng, and Jordan}]{blei2003latent}
David~M Blei, Andrew~Y Ng, and Michael~I Jordan. 2003.
\newblock Latent dirichlet allocation.
\newblock \emph{the Journal of machine Learning research}, 3:993--1022.

\bibitem[{Boyd-Graber et~al.(2017)Boyd-Graber, Hu, Mimno
  et~al.}]{boyd2017applications}
Jordan~L Boyd-Graber, Yuening Hu, David Mimno, et~al. 2017.
\newblock \emph{Applications of topic models}, volume~11.
\newblock Now Publishers Incorporated.

\bibitem[{Devlin et~al.(2019)Devlin, Chang, Lee, and
  Toutanova}]{devlin-etal-2019-bert}
Jacob Devlin, Ming-Wei Chang, Kenton Lee, and Kristina Toutanova. 2019.
\newblock \href {https://doi.org/10.18653/v1/N19-1423} {{BERT}: Pre-training of
  deep bidirectional transformers for language understanding}.
\newblock In \emph{Proceedings of the 2019 Conference of the North {A}merican
  Chapter of the Association for Computational Linguistics: Human Language
  Technologies, Volume 1 (Long and Short Papers)}, pages 4171--4186,
  Minneapolis, Minnesota. Association for Computational Linguistics.

\bibitem[{Gao et~al.(2021)Gao, Yao, and Chen}]{gao-etal-2021-simcse}
Tianyu Gao, Xingcheng Yao, and Danqi Chen. 2021.
\newblock \href {https://doi.org/10.18653/v1/2021.emnlp-main.552} {{S}im{CSE}:
  Simple contrastive learning of sentence embeddings}.
\newblock In \emph{Proceedings of the 2021 Conference on Empirical Methods in
  Natural Language Processing}, pages 6894--6910, Online and Punta Cana,
  Dominican Republic. Association for Computational Linguistics.

\bibitem[{Greene and Cunningham(2006)}]{greene2006practical}
Derek Greene and P{\'a}draig Cunningham. 2006.
\newblock Practical solutions to the problem of diagonal dominance in kernel
  document clustering.
\newblock In \emph{Proceedings of the 23rd international conference on Machine
  learning}, pages 377--384.

\bibitem[{Grootendorst(2020)}]{grootendorst2020bertopic}
Maarten Grootendorst. 2020.
\newblock \href {https://doi.org/10.5281/zenodo.5564211} {Bertopic: Leveraging
  bert and c-tf-idf to create easily interpretable topics.}

\bibitem[{Jin et~al.(2021)Jin, Zhao, Liu, Du, and
  Buntine}]{jin-etal-2021-neural}
Yuan Jin, He~Zhao, Ming Liu, Lan Du, and Wray Buntine. 2021.
\newblock \href {https://aclanthology.org/2021.emnlp-main.80} {Neural
  attention-aware hierarchical topic model}.
\newblock In \emph{Proceedings of the 2021 Conference on Empirical Methods in
  Natural Language Processing}, pages 1042--1052, Online and Punta Cana,
  Dominican Republic. Association for Computational Linguistics.

\bibitem[{Kingma and Welling(2013)}]{kingma2013auto}
Diederik~P Kingma and Max Welling. 2013.
\newblock Auto-encoding variational bayes.
\newblock \emph{arXiv preprint arXiv:1312.6114}.

\bibitem[{Lim and Buntine(2015)}]{pmlr-v39-lim14}
Kar~Wai Lim and Wray Buntine. 2015.
\newblock \href {https://proceedings.mlr.press/v39/lim14.html} {Bibliographic
  analysis with the citation network topic model}.
\newblock In \emph{Proceedings of the Sixth Asian Conference on Machine
  Learning}, volume~39 of \emph{Proceedings of Machine Learning Research},
  pages 142--158, Nha Trang City, Vietnam. PMLR.

\bibitem[{Liu et~al.(2019)Liu, Ott, Goyal, Du, Joshi, Chen, Levy, Lewis,
  Zettlemoyer, and Stoyanov}]{liu2019roberta}
Yinhan Liu, Myle Ott, Naman Goyal, Jingfei Du, Mandar Joshi, Danqi Chen, Omer
  Levy, Mike Lewis, Luke Zettlemoyer, and Veselin Stoyanov. 2019.
\newblock Roberta: A robustly optimized bert pretraining approach.
\newblock \emph{arXiv preprint arXiv:1907.11692}.

\bibitem[{McInnes and Healy(2017)}]{mcinnes2017accelerated}
Leland McInnes and John Healy. 2017.
\newblock Accelerated hierarchical density based clustering.
\newblock In \emph{Data Mining Workshops (ICDMW), 2017 IEEE International
  Conference on}, pages 33--42. IEEE.

\bibitem[{McInnes et~al.(2018)McInnes, Healy, and Melville}]{mcinnes2018umap}
Leland McInnes, John Healy, and James Melville. 2018.
\newblock Umap: Uniform manifold approximation and projection for dimension
  reduction.
\newblock \emph{arXiv preprint arXiv:1802.03426}.

\bibitem[{Miao et~al.(2016)Miao, Yu, and Blunsom}]{miao2016neural}
Yishu Miao, Lei Yu, and Phil Blunsom. 2016.
\newblock Neural variational inference for text processing.
\newblock In \emph{International conference on machine learning}, pages
  1727--1736. PMLR.

\bibitem[{Nan et~al.(2019)Nan, Ding, Nallapati, and
  Xiang}]{nan-etal-2019-topic}
Feng Nan, Ran Ding, Ramesh Nallapati, and Bing Xiang. 2019.
\newblock \href {https://doi.org/10.18653/v1/P19-1640} {Topic modeling with
  {W}asserstein autoencoders}.
\newblock In \emph{Proceedings of the 57th Annual Meeting of the Association
  for Computational Linguistics}, pages 6345--6381, Florence, Italy.
  Association for Computational Linguistics.

\bibitem[{Newman et~al.(2010)Newman, Lau, Grieser, and
  Baldwin}]{newman-etal-2010-automatic}
David Newman, Jey~Han Lau, Karl Grieser, and Timothy Baldwin. 2010.
\newblock \href {https://aclanthology.org/N10-1012} {Automatic evaluation of
  topic coherence}.
\newblock In \emph{Human Language Technologies: The 2010 Annual Conference of
  the North {A}merican Chapter of the Association for Computational
  Linguistics}, pages 100--108, Los Angeles, California. Association for
  Computational Linguistics.

\bibitem[{Ramos et~al.(2003)}]{ramos2003using}
Juan Ramos et~al. 2003.
\newblock Using tf-idf to determine word relevance in document queries.
\newblock In \emph{Proceedings of the first instructional conference on machine
  learning}, volume 242, pages 29--48. Citeseer.

\bibitem[{Reimers and Gurevych(2019)}]{reimers-gurevych-2019-sentence}
Nils Reimers and Iryna Gurevych. 2019.
\newblock \href {https://doi.org/10.18653/v1/D19-1410} {Sentence-{BERT}:
  Sentence embeddings using {S}iamese {BERT}-networks}.
\newblock In \emph{Proceedings of the 2019 Conference on Empirical Methods in
  Natural Language Processing and the 9th International Joint Conference on
  Natural Language Processing (EMNLP-IJCNLP)}, pages 3982--3992, Hong Kong,
  China. Association for Computational Linguistics.

\bibitem[{R{\"o}der et~al.(2015)R{\"o}der, Both, and
  Hinneburg}]{roder2015exploring}
Michael R{\"o}der, Andreas Both, and Alexander Hinneburg. 2015.
\newblock Exploring the space of topic coherence measures.
\newblock In \emph{Proceedings of the eighth ACM international conference on
  Web search and data mining}, pages 399--408.

\bibitem[{Sia et~al.(2020)Sia, Dalmia, and Mielke}]{sia-etal-2020-tired}
Suzanna Sia, Ayush Dalmia, and Sabrina~J. Mielke. 2020.
\newblock \href {https://doi.org/10.18653/v1/2020.emnlp-main.135} {Tired of
  topic models? clusters of pretrained word embeddings make for fast and good
  topics too!}
\newblock In \emph{Proceedings of the 2020 Conference on Empirical Methods in
  Natural Language Processing (EMNLP)}, pages 1728--1736, Online. Association
  for Computational Linguistics.

\bibitem[{Srivastava and Sutton(2017)}]{srivastava2017autoencoding}
Akash Srivastava and Charles Sutton. 2017.
\newblock Autoencoding variational inference for topic models.
\newblock \emph{arXiv preprint arXiv:1703.01488}.

\bibitem[{Terragni et~al.(2021)Terragni, Fersini, Galuzzi, Tropeano, and
  Candelieri}]{terragni-etal-2021-octis}
Silvia Terragni, Elisabetta Fersini, Bruno~Giovanni Galuzzi, Pietro Tropeano,
  and Antonio Candelieri. 2021.
\newblock \href {https://doi.org/10.18653/v1/2021.eacl-demos.31} {{OCTIS}:
  Comparing and optimizing topic models is simple!}
\newblock In \emph{Proceedings of the 16th Conference of the European Chapter
  of the Association for Computational Linguistics: System Demonstrations},
  pages 263--270, Online. Association for Computational Linguistics.

\bibitem[{Thompson and Mimno(2020)}]{thompson2020topic}
Laure Thompson and David Mimno. 2020.
\newblock Topic modeling with contextualized word representation clusters.
\newblock \emph{arXiv preprint arXiv:2010.12626}.

\bibitem[{Xia et~al.(2020)Xia, Wu, and Van~Durme}]{xia-etal-2020-bert}
Patrick Xia, Shijie Wu, and Benjamin Van~Durme. 2020.
\newblock \href {https://doi.org/10.18653/v1/2020.emnlp-main.608} {Which
  *{BERT}? {A} survey organizing contextualized encoders}.
\newblock In \emph{Proceedings of the 2020 Conference on Empirical Methods in
  Natural Language Processing (EMNLP)}, pages 7516--7533, Online. Association
  for Computational Linguistics.

\bibitem[{Zhao et~al.(2021)Zhao, Phung, Huynh, Jin, Du, and
  Buntine}]{zhao2021topic}
He~Zhao, Dinh Phung, Viet Huynh, Yuan Jin, Lan Du, and Wray Buntine. 2021.
\newblock Topic modelling meets deep neural networks: A survey.
\newblock \emph{arXiv preprint arXiv:2103.00498}.

\end{thebibliography}
\bibliographystyle{acl_natbib}

\clearpage
\onecolumn
\appendix

\section{Configuration Details}
\label{appendix_implementation_details}

We implement LDA and NTMs based on OCTIS \citep{terragni-etal-2021-octis} \footnote{\url{https://github.com/MIND-Lab/OCTIS}} and use their default settings. Specifically, ProdLDA, CombinedTM, and ZeroShotTM share the same configurations, i.e. one hidden layer with 100 neurons, ADAM optimizer and Momentum as 0.99; we randomly dropout 20\% hidden units; we run 100 epochs of each model, and the batch size is 64. For BERT+KM, we follow \citet{sia-etal-2020-tired} by reducing embedding dimension to 50 using Principal Component Analysis (PCA) and adopting TF to select words. For BERT+UMAP+HDBSCAN, we follow BERTopic \citet{grootendorst2020bertopic} and allows it to reduce the topic numbers. For our methods, we implement clustering-based experiments based on BERTopic \citep{grootendorst2020bertopic} \footnote{\url{https://github.com/MaartenGr/BERTopic}}. We reduce embedding dimension to 5 using UMAP. We use BERT, RoBERTa, and SBERT embeddings provided by HuggingFace \footnote{\url{https://huggingface.co/models}}, and SimCSE embeddings provided from its official Github \footnote{\url{https://github.com/princeton-nlp/SimCSE}}.

\section{Comparison of Topic Words}
\label{compare_topic_words}

We run Sup-SimCSE(RoBERTa$_{\texttt{base}}$)+UMAP on 20Newsgroup and show the differences of topic diversities produced by distinct word selecting methods in Table \ref{topic_words_comparison}. It is clear that $\mathbf{TFIDF_{i}}$ and $\mathbf{TF_{i}}$ tend to choose common words across multiple topics.

\begin{table*}[h]
\small
\centering
\begin{tabular}{c|c|l}

\hline
\multicolumn{1}{c|}{\textbf{Topic}} & \multicolumn{1}{c|}{\textbf{Weighting Method}}  & \multicolumn{1}{c}{\textbf{Topic Words}} \\
\hline
\multirow{3}{*}{Topic 1} & $\mathbf{TFIDF \times IDF_{i}}$ & car bike ride engine brake tire drive mile road front \\
                         & $\mathbf{TFIDF_{i}}$ & car bike \textbf{good} brake drive \uline{\textbf{make}} ride \uline{\textbf{time}} engine tire \\
                         & $\mathbf{TF_{i}}$ & car bike \textbf{good} drive \uline{\textbf{make}} \uline{\textbf{time}} engine ride back \uline{\textbf{year}} \\
\hline
\multirow{3}{*}{Topic 2} & $\mathbf{TFIDF \times IDF_{i}}$ & armenian turkish people kill israeli genocide village jewish war government \\
                         & $\mathbf{TFIDF_{i}}$ & armenian people turkish genocide government \uline{\textbf{make}} israeli kill \uline{\textbf{time}} village \\
                         & $\mathbf{TF_{i}}$ & people armenian turkish \uline{\textbf{make}} kill government \uline{\textbf{time}} \uline{\textbf{year}} state child \\
\hline
\end{tabular}

\caption{Comparison of topic words generated using different weighting methods when $K = 30$. Repeated words across topics are marked with an underline. Incoherent words are in bold. }
\label{topic_words_comparison}
\end{table*}





\end{document}